# Analysis of Deep Learning Architectures and Efficacy of Detecting Forest Fires

Ryan Marinelli

Submitted for the degree of Master of Science
University of York
Department of Computer Science
March 2022

THE UNIVERSITY *of York*

# Acknowledgements

I would like to thank my advisor Johnson Eze for providing guidance and managing the difficulties of research throughout this research initiative. I also would like to thank Mathis Lucka as my mentor in machine learning through his involvement with the Correlaid non-profit. Additionally, I would like to thank Robie Seeburger (Drexel University) for helping to grow my interest in computer vision and encouraging me to never settle.

# Table of Contents

LIST OF FIGURES

# 1. Executive Summary

The aim of this research is to review the state of computer vision as applied to combatting forest fires. My motivation to research this topic comes from the urgency with which new participants and stakeholders require guidance in this field. One of these new stakeholder groups are practitioners of machine learning that lack domain expertise. Introducing these new entrants to domain specific datasets and methods is critical to supporting this aim as general computer vision datasets are insufficient to support specialized research initiatives [1]. The overarching aim of the research is to introduce datasets and methods to make them more accessible to the community.

There are two main architectures that are reviewed in this work: Convolutional Neural Networks (CNN) and transformers. Given that research is taking place in an Internet of Things environment, how these models respond to optimizing is significant. Pruning and quantization are two popular optimization techniques that allow for an ease of computation while sacrificing accuracy. This trade-off and how to navigate it is at the core of this endeavor. This trade-off is evaluated by training two popular variations of CNN and a transformer on three keystone datasets. These architectures are evaluated in terms of accuracy and precision with other key metrics. Pruning and then quantization is applied per model per evaluation. The goal is to determine how accuracy is comprised per optimization applied. It appears that transformers perform more highly overall, but they are more affected by pruning. In the worst case, a transformer lost 34% accuracy after being pruned. A CNN in the same circumstances gained 13% accuracy as pruning removed superfluous complexity. However, quantization undermines CNN. These different reactions create space for the practitioners to navigate optimization decisions. As for other considerations, there is also no ethical or social dimension to the work as it targets more applied problems in the field.

# 2. Introduction

The aim of this research endeavor is to explore the state of computer vision and how it may be a tool in combating climate change. One avenue for leveraging technology is to improve reaction time to wildfires. Forest and wildfires have been selected as the focus since they provide a unique opportunity for mitigation.

## 2.1 Motivating the Problem

The risk forest fires pose is consistently increasing as a component of the emerging climate crisis. From 2011 to 2020 in the US, 62,805 wildfires burned on average with a total impact of 7.5 million acres impacted annually [2].

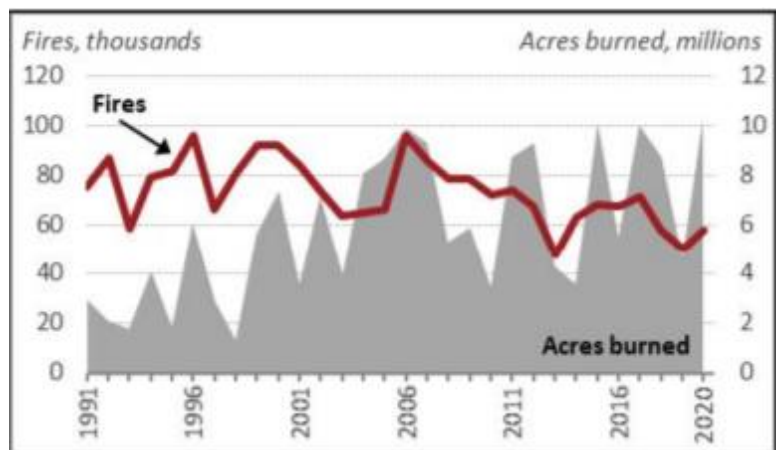

Figure 1: Data Concerning Proliferation of Fires in US

Source:  NICC Wildland Fire Summary and Statistics annual report

It is apparent that this problem is continuing to grow in scope and is increasingly growing costly to manage. In terms of total cost, it has been estimated the wildfires in California cost the state $148.5 billion USD in 2018 [4]. Given this damage, there is significant motivation at attempting to curb the destruction wildfires are causing. The approach that is being embraced is to use of UAV (Unmanned Aerial Vehicles) to monitor areas and to facilitate intervention.

In current usage, there are 65 drone operators that are servicing firefighting initiatives through the California Department of Forestry and Fire Protection [5]. Most of the initiatives with UAV's concern reconnaissance and monitoring fires. They have been used in circumstances where it is too dangerous to fly a piloted vehicle and are starting to focus on taking a more proactive stance in terms of managing fires.  With more technology and detection methods, it may be possible to enhance the capabilities of drones to mitigate and manage fires more effectively.

## 2.2 Outlining AI and Pathway to Achieve Aims

In general, there are two main varieties of computer vision models. The first are those made with the intention of object detection. The goal of this variety is to determine if there is an object in an image and to assign coordinates based on where the object is in a frame and to superimpose a bounding box on the image. One of the most popular models for this use case is YOLO (You Only Look Once). This model is reviewed in this paper as the current state of artificial intelligence. It makes sense as an industry standard and is a good benchmark.

The second variety of computer vision uses image classification. The goal of these models is to assign a class to an image. In this context, photos are labeled in a binary fashion. They are typically "fire" or "not fire" in the literature. The first images contain forest fires as a positive class. The other contains a negative class of images that are likely to confuse the model. For instance, morning steam is important to include in the data. As the visual qualities of steam are analogous with smoke, it has been the cause of many false positives. Thus, the dataset is just as significant in developing effective models as architectures are in performance. One of the highest quality datasets for wildfire detection is the FLAME dataset from Shamsoshoara et al. It contains footage from drones and includes data from thermal cameras. The FLAME dataset is reviewed in this research for its quality. The associated model from Shamsoshoara et al. is also reviewed. It is tuned for the specifics of aerial image classifications and contrasts well to YOLO. As YOLO is more general, it is useful to study as it explores variation between specialized and generalized approaches. Additionally, newer architectures are reviewed to determine marginal gains within the fire monitoring domain. Specifically, transformer architectures have gained prominence recently and represent the "future state" of detection mechanisms. As for considerations per the use case, optimizations for models and their deployment will be evaluated to determine how they affect performance. Some of the popular techniques applied are pruning and quantization and are meant to filter out non-essential elements of the model and are designed to lift burdens in terms of computational intensity. This is useful to consider given that edge processing is limited, and the environment is more constrained.

## 2.3 Research Contribution

Given this trade-off, the underlying research question is to measure how much loss is expected when performing computational optimalizations. To pursue this question, different models of interest are trained on keystone datasets to determine drops in accuracy and other key metrics. 28 models are trained and evaluated in this endeavor. The set of models trained constitutes three architectures of interest, on three datasets of interest, while having the models pruned and quantized. Ensemble models are also explored to evaluate model synergies. Through comparing the different iterations of models, it is possible to derive guidance how to approach fire detection through observing relative drops in accuracy and precision with the different models available to practitioners. In general, it is found that CNN models are easier to optimize than transformers. CNN models take only marginal penalties for pruning. However, they are undermined when quantization is applied. Transformers suffer more from pruning, but they are more robust when quantized. This creates a dichotomy to navigate for practitioners in determining the degree of optimization. This work provides guidance based on experimentation to provide guidance to better inform these decisions.

# 3.Literature Review

The results of machine learning are determined by two factors: the data and the architecture. The adage "garbage-in garbage out" defines the relationship with data and the underlying architecture. The quality of the data is what powers the model to learn. If there is a feature that is not present in the data or is overrepresented, then the model will be mis calibrated and will be evident in performance metrics. Thus, data is core to understanding how to best approach solving applied problems. Additionally, model architectures are as significant. The flexibility of a model to leverage the data and learn from the feature is the other central component to designing robust intelligent systems. An over-specified model may overfit on certain characteristics of the data and may be as problematic as having insufficient data. Both data and architectures require careful inspection to best address tackling applied machine learning problems to navigate potential hazards.

## 3.1 Data

Most computer vision research recycles the same benchmark datasets. Two of the most common datasets are Imagenet and Coco [6]. The problem with using these datasets for a particular project is that they are too generalized [1]. There are 1000 classes that are being classified. While this may be useful for certain use-cases, the lack of specialized datasets in fire science is a limitation in the field. Thus, a review of the most specialized datasets is significant to communicate what is available and to understand the strengths and shortcomings of the data.

### 3.1.1 BowFire

The BowFire (Best of Both Worlds Fire Detection) dataset comes from Chino et al. The goal of their research is to address high-false positive results [7]. Chino et al. was written in 2015 and is from a period in evolution in the literature. The problem formulation changed. Previous methods to identify fire were based on rules-based approaches using RGB/HSK values. The usage of image segmentation and CNN's would overtake these approaches with the approach in this paper contributing to the shift. While the methods are outmoded, the data in this paper is high quality. It was curated to have a variety of colors in the image set with a high amount of variation in image resolution with high variance in the contours of the flames.

There are 226 images in the dataset overall. There are 119 images with fire and 107 without. This provides a balanced dataset. There are also "fire-like" images. This includes images of the sunsets and other situations that may confuse algorithms. While the dataset may be lacking in terms of the quantity needed to train deep learning approaches, it offers significant variance that is of value to machine learning approaches generally.

### 3.1.2 FLAME

The FLAME (Fire Luminosity Airborne-based Machine learning Evaluation) provides aerial images of controlled burns. The data is used in Shamsoshoara Et. Al. The images are from a controlled burn using drones. There were three cameras used in the collection of the data: a Zenmuse X4S, Phantom 3 Camera, and a FLIR Vue Pro R. The FLIR camera is particularly interesting, since it captures infrared images. The drones used are DJI Matrice 200 and DJI Phantom 3 Professional. DJI products are rather standard and are quality; this also contributes to the value of the data.

This data is unique in that there are no other aerial images for fire analysis [8]. When gathering the data, there were different perspectives taken with different drones from a top down perspective. This viewpoint is significantly different than with a ground camera and provides more value to the dataset [8]. Given the uniqueness and quality of the data, it is ideal for the use case of developing models to detect fires.

The main weakness with the data is that it could allow for models created from it to be susceptible to concept drift [9]. Since the data is collected in January with snow on the ground, models would not be exposed to other coloration associated with seasonality. For instance, autumn leaves with bright colors and jagged contours could be a problem with different accuracy throughout the year. Collecting the data in winter exacerbates the issue in that a blanket of snow offers the greatest contrast to a blanket of colorful leaves.

### 3.1.3 Zhang Et. Al Dataset

The goal of Zhang Et Al is to create a mixed dataset that is large enough to be supportive of CNN to support the identification of smoke. The approach to generating the data is two-fold. A bush burn is captured behind a green screen, and the smoke bloom is cropped-out and superimposed on the background of a rural area in various positions to provide training data. Additionally, the researchers used Blender (a 3D modeling software) to create artificial smoke and to superimpose the artificial smoke in the same backdrop. 12,620 images were created using this method for training data. The data generated with Blender is seen as more precise and yielded better results. However, the data seems to provide insufficient coverage of thinner varieties of smoke and is a weakness the authors recognize [10]. Despite this shortcoming, the dataset provides an accessible dataset that can be trained to identify smoke and is a complementary dataset.

# 3.2 Model Architectures

## 3.2.1 YOLO

YOLO is considered the current state of the art in object detection and is widely used in industry. It historically has used Darknet as its basis. Darknet is a framework written in C for deep learning [11]. It was created before Tensorflow and Pytorch were developed. CSPDarknet53 is the backbone of YOLO [12]. It is a Cross-Stage Partial Network whose main achievement is reducing computation time allowing for depth-wise convolution networks to be computationally manageable [13]. This is accomplished by partitioning the base feature map and merging the partitioning within a hierarchy. The underlying objective is to distribute gradient propagation amongst different paths within a network to remove bottlenecks and increase the flow throughput [14].

YOLO uses PANet (Path aggregation network) as the neck of YOLO [15]. This network is designed to boost information flow for segmentation tasks. There are two aspects of the network that are novel. The first is using bottom-up path augmentation. This supports localization by propagating lower layer feature information throughout the network [15]. The other is adaptive feature pooling. This process links features with features from different levels within sub-networks and supports stronger representations of multi-level features [15]. Overall, it seeks to amplify localization to interconnect the lower layers to the feature map.

The head of YOLO produces three feature maps that are then applied to CSPDarkent53 and PANeT for feature fusion [13]. These three parts of YOLO encapsulate the core of the architecture and represent the most used model in production.

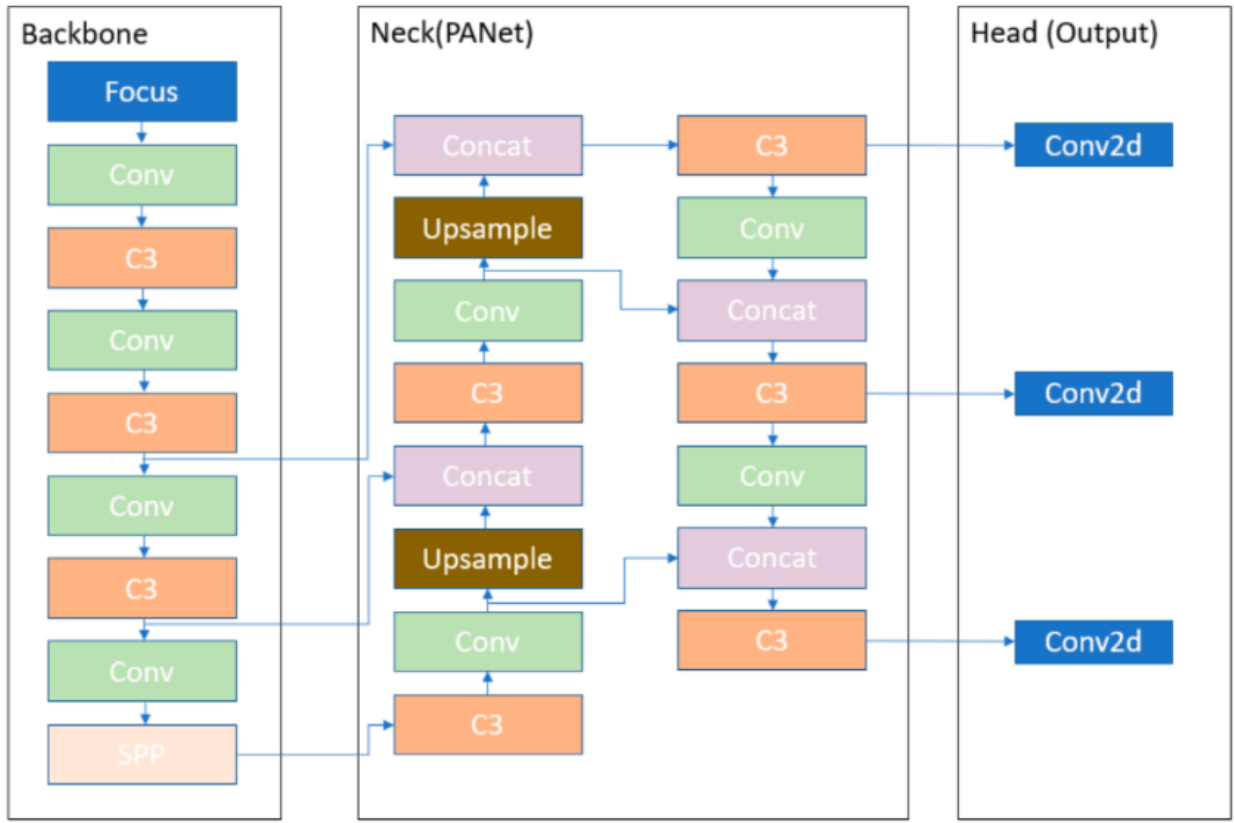

Figure 2: YOLO Architecture Overview From U. Nepal & H. Eslamiat

### 3.2.2 Xception

Xception is used in Shamsoshoara Et. al. Xception is a variety of CNN. In a convolutional layer, a CNN tries to map features from a 3-D space using length and width and a color channel. This requires the convolution to map both spatial correlations and correlations from the color channel [16]. Inception modules use these correlations and feed them into a smaller convolutional layer to map to a smaller 3-D representation [16]. The assumption that underlies the architectures is that cross-channel correlations and spatial correlations are not strongly correlated and that it is more effective to map the two spaces separately than jointly [16]. This hypothesis is at the core of Xception CNN. The idea is to have a disjoint mapping process between the cross-channel correlations and the spatial correlations and requires that intermediate convolutions be fed into a singular 1x1 convolution.

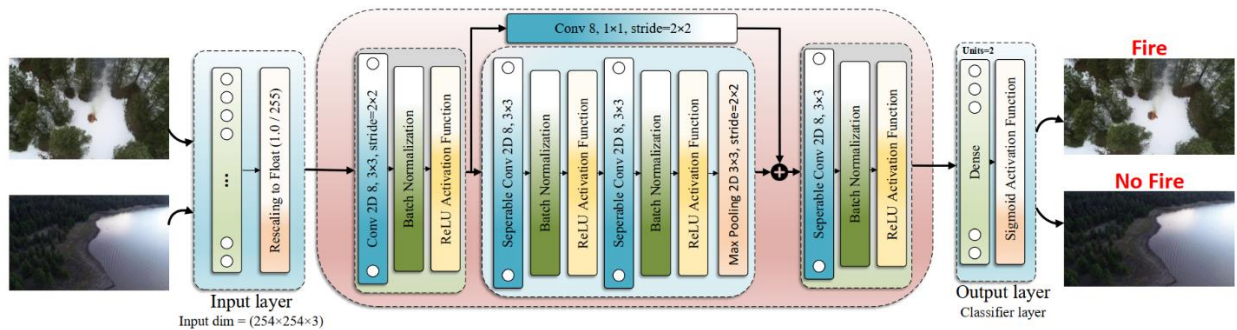

Figure 3: Architecture From Shamsoshoara Et. al.

The Xception architecture effectively uses a series of Inception modules in a depth-wise configuration. Traditionally, there are 36 layers for feature extraction and are organized into 14 modules into a linear stack with residual connections [16].

The distributive process through separating correlations appears analogous to the

segmentation process that occurs with CSPDarknet53. With Xception, the concern is more on

the representation of the feature space as it is more critical to the architecture. In

CSPDarket53, the concern is more computational in nature and emphasizes learning features through propagation flow rather than the representation of the features throughout the convolutional layers.

## 3.2.3 Vision Transformer

Transformers are an architecture that has been recently adapted from natural language processing to computer vision. Transformers split an image into patches and have a linear embedding applied to treat them as tokens. A position embedding is also applied, and the tokens are fed into an encoder [17].

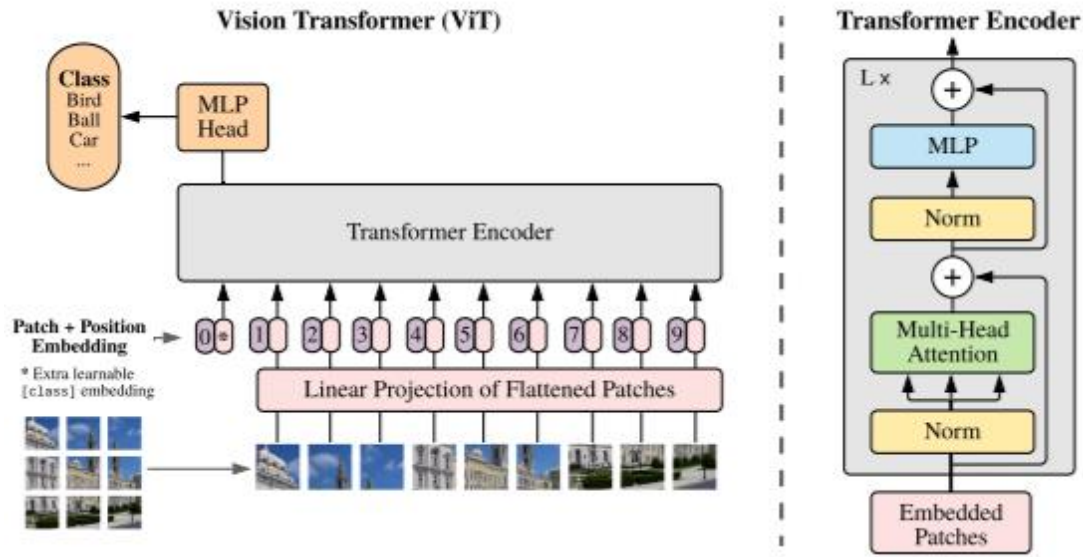

Figure 4:Overview of Transformer Architecture From Kolesnikov et al.

The main difference with this approach is the relationship of locality and how it maps to feature maps. CNN architectures can leverage the relationship of the convolutions for their input space. This is especially true for ResNets that are more explicit about the relationship. However, transformers can exceed CNN architectures given there is sufficient data to overcome inductive bias and the strengthened associated locality [17]. Thus, when training a transformer, it is usually necessary to apply transfer learning with a more specialized dataset to get optimal performance from a model.

As an emerging architecture, transformers are starting to be evaluated. In Shahid and Hua, a family of visual transformer models were trained on the BowFire dataset. The authors report 96% recall of using VIT/16 model using ImageNet weights [18]. This is the same process adhered to in this endeavor, and 95% recall is achieved. Thus, the research in this current endeavor is helping to build a sense of consensus and validation on the gains that transformers provide compared to other approaches.

The main weakness in their research is the data selection for their benchmarking. Unfortunately, the authors did not review aerial datasets like the FLAME and may have overstated claims. The data provided by aerial imagery is fundamentally different from what is provided in more conventional fire datasets. For instance, the flames are smaller with more environmental factors at play. The research initiative here seeks to fill in gaps in the literature by applying state of the art methods to the more realistic data.

### 3.2.4 Model Discussion

Given these models, there are a few points to consider that are driving architectures. The first is how to best address locality. Locality is essentially how to represent the relationship of segments of an input in a more complex feature space. CNN use convolutions as the main mechanism. This is seen in both YOLO through PathNet and the use of residuals in Xception. Vision Transformers are particularly weak in learning to represent locality and essentially have to brute force representations through more data [17]. It is of great interest to determine how locality may play a role in detecting fires. If fires are rather small and limited to patches, then the transformer may underperform. For object detection, locality is more important than in image classification and may pose additional problems. The research endeavor here is in part to review the models and to determine how applicable transformer methods are when classifying images of fire to guide future work in the field.

## 3.3. Model Optimizations

In the previous discussion, the emphasis has been on data and the architectures. The data is specialized to fit the needs of aerial fire detection. However, the architectures discussed remain broad. How can they be tailored to fit the needs of the particular use case? The goal of reviewing optimizations is to create leaner models that are more amiable for deployment in constrained computing environments.

### 3.3.1 Pruning

"Deep Compression: Compressing Deep Neural Networks With Pruning, Trained Quantization, and Huffman Coding" by Han et al describes the current state of the art methods for compressing models and making them deployable for an embedded systems use case. In CNN architectures, pruning should not cause a drop in accuracy and may reduce overfitting in models [19]. To prune models, the authors train a model under normal circumstances. Then, the smallest weights in the model are pruned away. The idea being some were spurious. The authors report that they reduced the number of parameters by 13 times with no accuracy loss in their VGG-16 model: a former state of the art CNN architecture.

The only weakness to the address in this paper is that it concentrates on reviewing CNN architectures as they were only state of the art at the time it was written. Transformers and their behavior under pruning and quantization were not discussed leaving another opportunity to address gaps in the literature.

### 3.3.2 Quantization

Quantization is also applied in-tandem with pruning in Han et al. By combining both methods, their VGC-16 models achieved 27 times compression. In terms of quantization, the general approach in the literature is to apply a linear transformation after training. The goal is to cast the weight from a float to a more discrete representation [20]. The underlying intuition of the optimization is that the value right of the decimal for a weight is not as significant as the value to the left. It takes significant memory to store all digits associated with smaller values that do not contribute as much to performance [20]. By remapping this data, the integrity of the weights is maintained without allocating the same degree of memory.

### 3.3.3 Optimization Discussion

Choosing to apply these optimizations may come at a cost, and it is significant to consider the trade-offs. For pruning models, some models may be resilient to losing accuracy. For instance, CNN seem to suffer little accuracy loss. This may not be the case for transformers. As for quantization, the loss is more recognized for greater gains in memory utilization and processing speed. Being explicit about these costs allows for more effective deployment of edge devices and is motivating analysis benchmarking.

# 4. Methodology

This analysis is at the intersection of two varieties of studies.  In essence, it is a case study. The analysis observes a particular narrowly defined domain and is attempting to optimize for the context of interest. Yet, in another sense, the analysis is a benchmarking study. Comparisons are being made across various metrics to determine generalizable truths about performance. When considering these two varieties of studies, they appear to be in contradiction. One is designed to be an in-depth, and the other is supposed to be generalizable. However, they do not contradict. They are synergistic approaches of analysis.

By emphasizing a particular domain there is a bound that may be imposed in a case study. This allows for the proper investigation of a phenomenon of interest [21]. The strength of a case-centric approach is trying to derive exploratory insights from a small N study. Given the nature of benchmarking, it is not feasible to make strong claims with statistical inference when investigating a phenomenon. Thus, the analysis will be mostly descriptive in nature and will be able to leverage the strength of a case-centric approach [21].  By imposing bounds within a constrained context, it possible to properly probe the particulars of the case and prioritize internal validity. In this context, internal validity is likely of greater value given the research is serving a particular group of stakeholders.

Additionally, this research is attempting to create consensus on how to approach optimization and modeling. Benchmarking is effective at paradigm building and taking an abstract concept and operationalizing [22]. By training different permutations of models and evaluating them, it makes theoretical trade-offs more concrete. This is made possible in part through the selection of data and context. By emphasizing tailored datasets within a specific domain, the results are more pertinent to the audience.

# 4.1 Data Preparation

Given the significance of data, it is of importance to discuss the pre-processing steps necessary to prepare the models. Each of the datasets were prepared with a split with validation. 80% of the data is used training the model and 20% is used in validation. Additional steps were taken depending on the model applied.

## 4.1.1 YOLO

YOLO requires annotated data following a specified file structure. The highest-level directories are label and images. Each have a "train" and "val" subdirectory. The label files contain the positions of the points to define a bounding box per class [23]. The tool DarkMark is used for labeling the data. It is part of the darknet toolchain and generates configuration files [24]. Due to the burden of annotating data, 800 images were annotated from the FLAME dataset and 500 images were annotated from the smoke dataset from Zhang et.al.

## 4.1.2 Xception

There are two directories in the data used to train Xception as part of the FLAME dataset. It is split into a train and test set with a "fire" and "no fire" directory within each of higher directories. Keras is used to apply data augmentation to the data set. A horizontal flip is used with an additional random rotation to increase the amount of variance found in the dataset. Augmenting data is suggested when training CNN as a means of avoiding overfitting and regularizing inputs [25]. Other augmentations were avoided due to potential concerns with the model learning spurious color gradients.

### 4.1.3 Visual Transformer

The files were compiled into two directories with a prefix added to the file names. These prefixes designated the class of the files and were labeled accordingly. A generator is used to apply augmentations and to create a balanced training, validation, and test set. Training balanced datasets is recommended when training classifiers [26]. The goal of using generators in this fashion is to give the transformer an opportunity to learn localized features to overcome potential weaknesses in the architecture to better discriminate between classes.

## 4.2 Training

The models are trained in two different environments. YOLO is trained in Google Colab to take advantage of their GPU. The time to upload the other datasets made working with Colab prohibitive. They were trained with an 8th generation I7 with an external SSD to increase the read speed for training. Other cloud-based solutions were considered, but the network speed was problematic.

### 4.2.1 YOLO

YOLO is trained using YOLOv5s. This is a point of considerable consideration. Originally, YOLOv4 was targeted as the basis for the comparison. There is a debate in the community over the legitimacy of YOLOv5 as an extension of YOLO [27]. In part because it was released without a white paper with less rigorous benchmarking.

However, that was at the launch of YOLOv5. There have been six versions adding credibility to the model and performance. It is also the most widely adopted and supported version of YOLO and is most likely to be used by the community. Some effort went into developing modeling in YOLOv4, but the ecosystem is written in C and requires using make files and is a more complex ecosystem with lesser adoption.

YOLOv5s is the version of YOLOv5 designed for more constrained environments. The "s" designates that it is small and is a lightweight implementation. There is a nano version, but it seemed to be too significant of a trade-off.

The Neural Magic API is leveraged in this endeavor. Neural Magic maintains a "model zoo" of sparse models for deployment that directly integrates into YOLO. The quantized YOLOv5s model and pruned YOLOv5s models were used [28]. For data visualization, Weights and Bias is used. This tool integrates into YOLO and is part of YOLO's documentation [29]. It provides interactive logs for experimentation.

## 4.2.2 Xception

The model is trained by running a script provided by the author. Tensorflow and Keras are used to train and prepare the model. The Tensorflow optimization library is used to perform pruning of the dense layers of the model. Tensorflow's documentation is adhered to while perform pruning [29]. 8-Bit quantization is applied to the dense layers of the model, and the documentation is adhered to in a similar fashion [31]. It essentially converts the float data to ease computational concerns.

## 4.2.3 Visual Transformer

This model is trained using a Keras implementation with the vit-keras package [32]. This is an implementation based on the Google Research team and recommends the use of transfer learning. The main difference when training the transformers is that ImageNet weights were used as the base and were then tuned with their respective datasets. This is significant in that a larger dataset is needed to support transformers given their weaknesses in locality as aforementioned. Pruning is applied to the dense layers in the tail of the model. Quantization is applied on the dense layers of the pruned model. The training process uses a generator to create augmented data that is balanced and follows recommended implementations of ViT [33]. The main drawback of using more prepackaged implementations is that it makes it difficult to optimize. Since the ViT model is read in as a subclass of Keras models, the Tensorflow API did not identify dense layers to prune or quantize in the head, so only layers in the tail had optimization methods applied.

## 4.3 Evaluation

A validation dataset is used to compare the results of the models. It comprises 20% of the relevant data. Precision and recall are the main metrics considered in this analysis. Recall is the ability of a model to correctly identify the object being detected. It is defined by the ratio of true positives to the ground truth [33] The ground truth being the total of true positives and false negatives. Precision is defined by the ability to recognize true positives. It is the ratio of true positive to true positive and false positives [34]. The main difference between these two metrics is that false negatives are used in recall while false positives are used with precision. These metrics are usually viewed together to understand the trade-offs a model is making in classification. Average precision is a metric that finds the area of the curve associated with this trade-off [28]. It attempts to find a more generalizable metric.

### 4.3.1 YOLO

For object detection, mean average precision is used. mAP uses "Intersection over the Union" when making determinations with false positives and true negatives. It is essentially a ratio of the overlap of labels and if they are correctly identified. It provides the most information on the general performance of the model [34]. Recall and precision curves are also reviewed to get an understanding of the trade-offs required when making optimizations.

### 4.3.2 Xception & Visual Transformer

To review the success of the model, a confusion matrix is created to determine the recall and precision of the model per optimization applied. This is more straightforward than in object detection. Both models use similar formulation in that they both address binary classification. The confusion matrix allows for a framework comparing true and false positives and negatives. The only problem is that due to the balanced nature of the training and test data for the transformer, true positives were equal to false positives and gave consistent metric results throughout. Thus, accuracy is reported but not the other metrics as they were consistent. While this allows the transformer to be balanced in terms of recall and precision, it makes it lack specialization and could be a limitation in this endeavor.

## 4.4 Additional Methodological Discussion

When considering decisions in terms of methodology, the goal is to adhere to the audience of this research: applied machine learning practitioners. Practitioners want to use well-supported tools and toolchains to support iteration. This is in part why Weights & Biases is used for visualization, why YOLOv5 is used, and why a package is used to implement ViT.

# 5. Analysis of Model Performance

The models will be compared using recall, precision, and accuracy for each of the varieties of models. One of the most informative comparisons is to observe the relationship between recall and precision that share a dichotomous relationship. This requires the navigation of the trade-off.

## 5.1 YOLO Evaluation

To build an intuition of the model, it is useful to do an inspection of a mosaic of the results for object detection. This provides useful feedback on what the model may conflate and the strengths the model possesses before reviewing quantitative metrics.

### 5.1.1 BowFire Dataset

In this mosaic, there are two classes of images. There are images with fire and without fire.

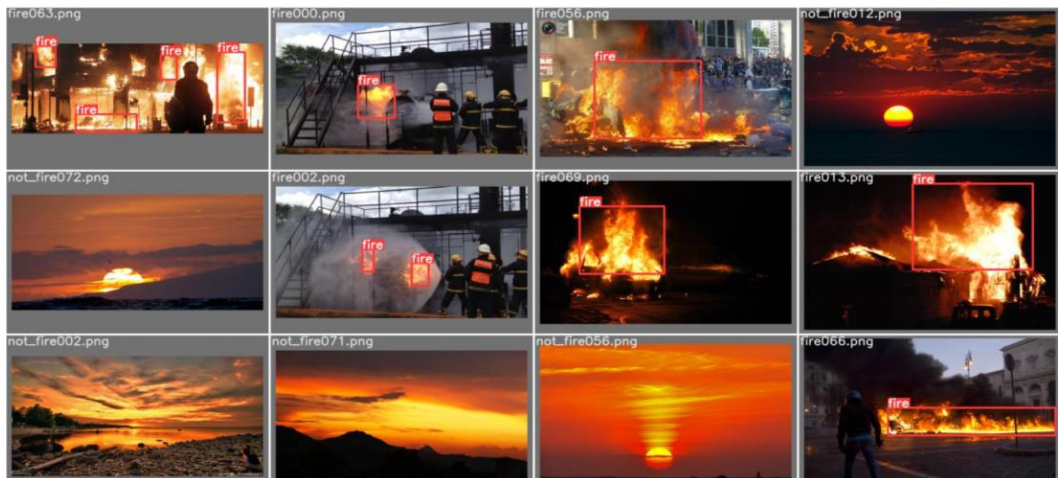

Figure 5:Mosaic of Object Detection Using YOLO with BowFire

The images without fire are significant in that it teaches the model to avoid common pitfalls. In particular, the sun and the reflection of it off water are historically problematic. The intuition is that the ripple of the water with the color could be associated with the contours of a flame.  It appears that YOLO is effectively able to discern between fire and common pitfalls suggesting that it is particularly robust. However, it does seem to miss some fire in images where it is rather abundant. For instance, in "fire063", it does not label all the instances.

## 5.1.1.1 Base Model

| Comparison of YOLO With BowFire | | | |
|---|---|---|---|
| **Models** | **Precision** | **Recall** | **mAP** |
| Base Model | 100% | 100% | 91% |
| Pruned Model | 38% | 41% | 7% |
| Quantized Model | 52% | 55% | 18% |
| Reports max value assiociated | | | |

Figure 6:Comparison of YOLO With BowFire

The basic version of YOLO seems to do generally well on this dataset. It reaches a mAP of 91%. The concern is that it is probably being overfit by having too many epochs applied. This is in relation to the relative size of the dataset. It is roughly only 150 images, while the standard is usually around 1000 images per class. The precision and recall reach a max value of 1.

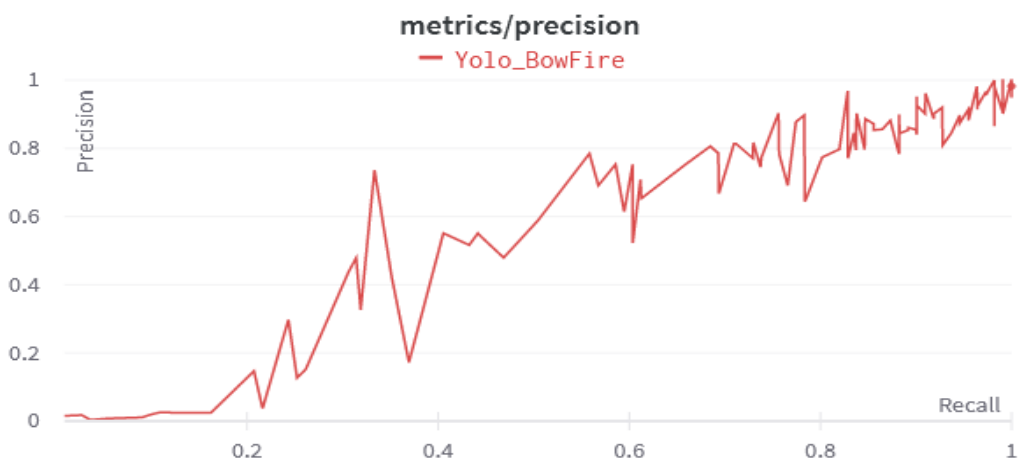

Figure 7:Recall/Precision for YOLO Base with BowFire

When observing the trade-off between recall and precision, the model appears to favor precision over recall. This is noted in earlier training epochs before the model becomes progressively overfit. This suggests the model is relatively more effective at avoiding false positives than negatives.

## 5.1.1.2 Pruned Model

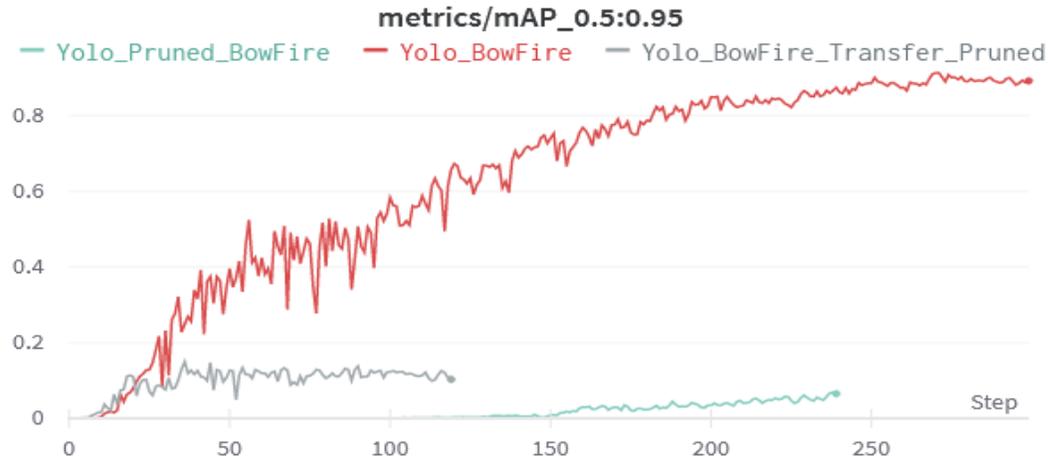

Figure 8:Comparison of mAP between YOLO Base and YOLO Pruned with BowFire

When pruned, the model archives 7% mAP. Considering the data available, it seemed prudent to compare how using transfer learning to fine tune the model would perform as well. Coco weights were used in this process. It reached around 10% mAP. The aspect of this comparison that is so stark is the difference between performance between the base model and the pruned version.

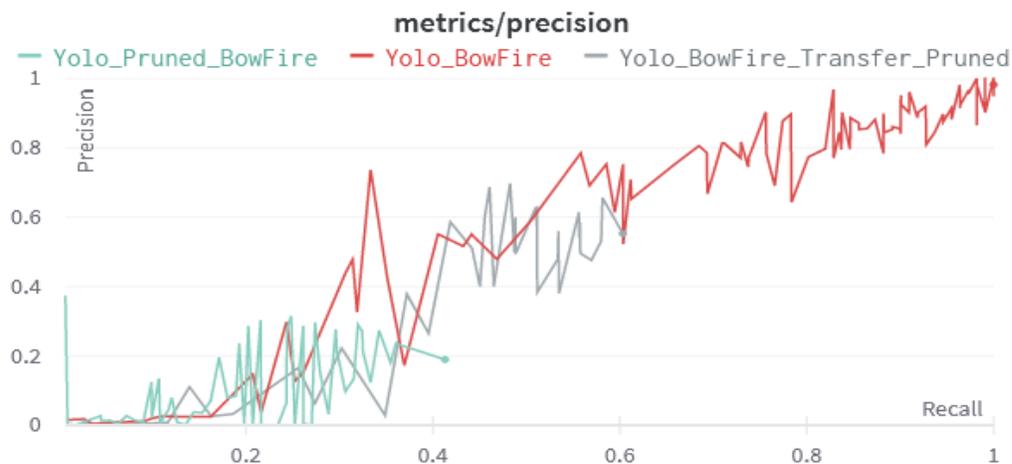

Figure 9:Recall/Precision with YOLO Base and YOLO Pruned with BowFire

When looking at the relationship between recall and precision, the model through transfer learning is far more competitive to the base YOLO model. The pruned YOLO model trained only on the Bowfire dataset clearly underperforms. This strongly suggests that transfer learning is advisable when working with limited datasets and to avoid penalties.

### 5.1.1.3 Quantized Model

The quantized version of the model achieves 18% mAP. The interpretation of the difference between the quantized model and the pruned model is that the pruned model is likely pruning on overfit weights. When significant weights were pruned away, it may have destroyed the performance of the model. When quantizing the model, it may have discarded some of the influences that overfit the model and increased performance. This is likely attributable to the smaller size of the dataset. These characteristics are likely not present in larger datasets as transfer learned provided some dampening.

The model achieves 55% recall and 52% precision. Thus, the quantized model is outperforming the pruned version.

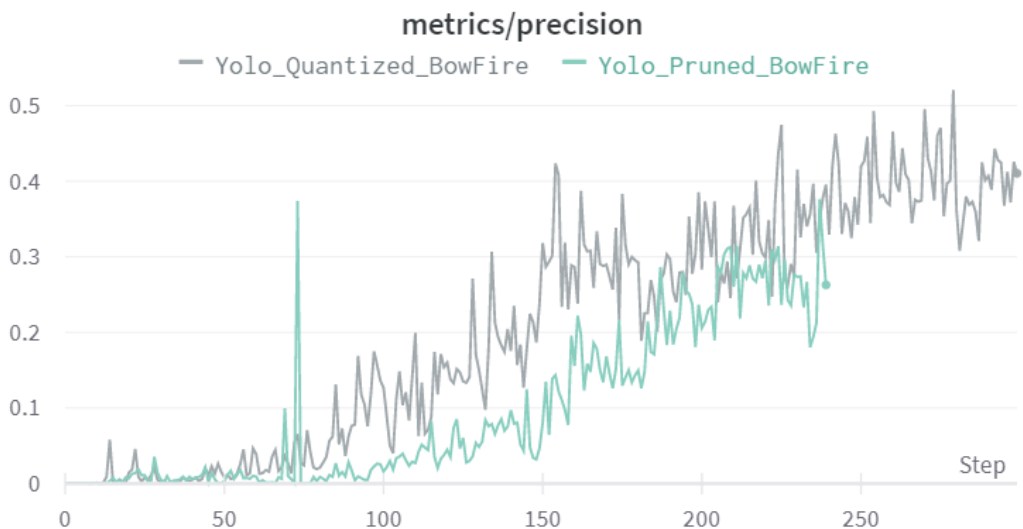

Figure 10: Recall/Precision with YOLO Base and Quantized with BowFire

## 5.1.2 FLAME Dataset

The FLAME dataset is significantly more difficult to detect fire from. This is essentially that the data approximates what a drone would see at the start of a forest fire. These flames are relatively small and are taken from an aerial perspective.

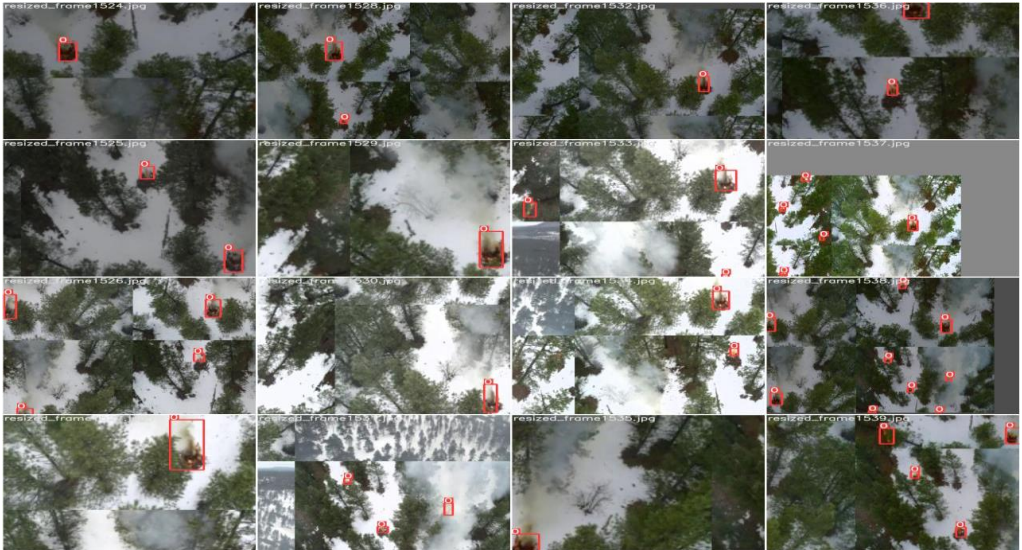

Figure 11:Mosaic of YOLO Fire Detection with FLAME dataset

In terms of detection, it does appear to do well. It is able to detect the controlled burn set by researchers. It is also able to detect that it is starting to spread to the branches of nearby trees in some circumstances. The most significant weakness is that it is not making use of the smoke in terms of the detection.

## 5.1.2.1 Base Model

The model achieves 18% mAP. While this is less performant when compared to BowFire, there is an important consideration to make. In BowFire, the bounding boxes were rather large. The fires were rather prominent in the images. In this dataset, the images are aerial in nature. The bounding boxes are significantly tighter and affects the IOU. Thus, the model should not be held to the same standard in terms of metrics.

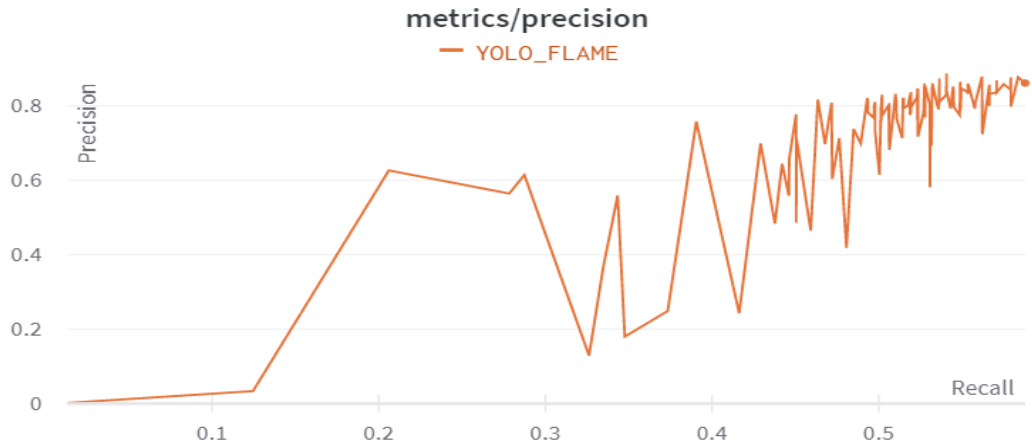

Figure 12: Recall/Precision of YOLO Base with FLAME

The model seems to perform well balancing precision and recall overall. There are areas of the graph where the model seems to favor precision. For instance, when the model reaches 70% precision, the associated recall is 40%.

| Comparison of YOLO With FLAME | | | |
| --- | --- | --- | --- |
| Models | Precision | Recall | mAP |
| Base Model | 92.14% | 59% | 18.49% |
| Pruned Model | 91.22% | 74% | 25.47% |
| Quantized Model | 88.76% | 85% | 24.48% |
| Reports max value assiociated | | | |

Figure 13:Comparison of YOLO With FLAME

## 5.1.2.2 Pruned Model

When pruned, the model reaches 25% mAP. The increase in performance may be in part due to the association that pruning alleviates overfitting [19]. Given that roughly 800 images were annotated for training, it would be sensible to see a performance boost in smaller datasets.

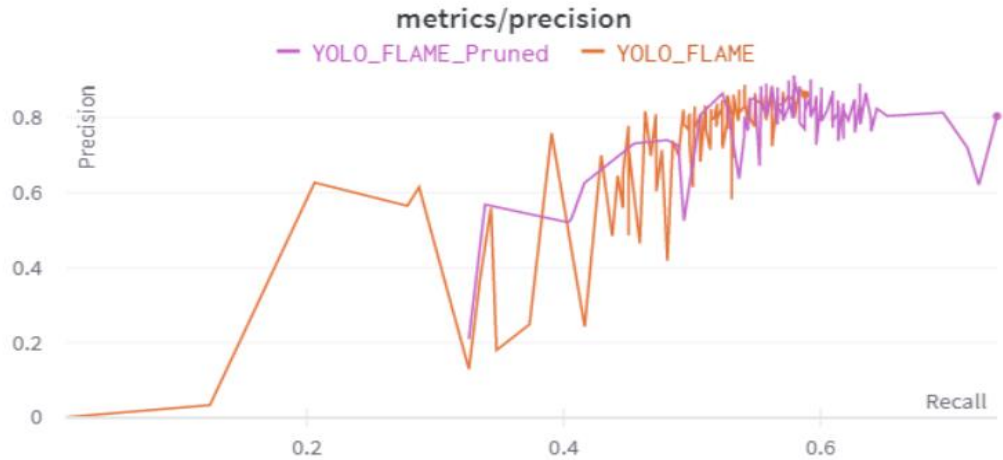

Figure 14: Recall/Precision of YOLO Base and Pruned with FLAME

The models are generally comparable. The base version of the mode appears to have higher precision than the pruned version. It favors recall and is more receptive to avoiding false negatives compared to the base model.

### 5.1.2.3 Quantized Model

The quantized version of the model reaches 24% mAP, so it slightly underperforms the pruned version in this respect. Otherwise, the model is rather comparable.

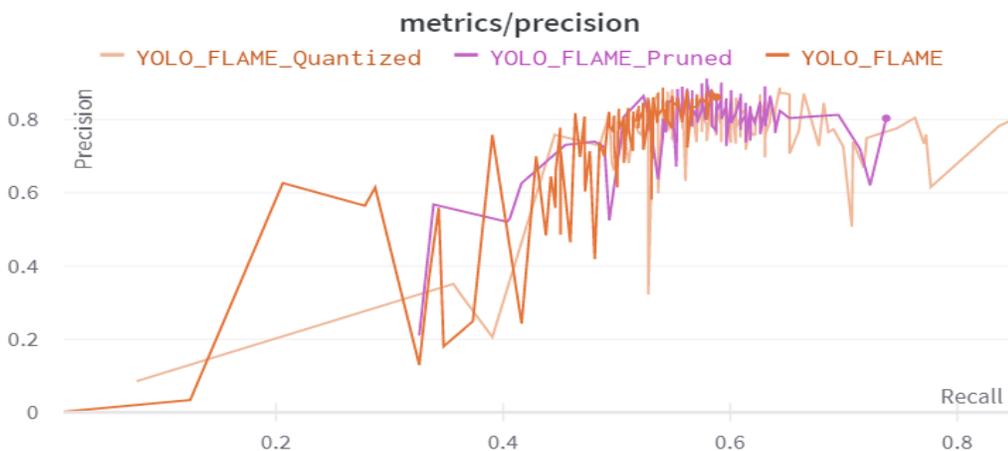

Figure 15:Recall/Precision of YOLO across models with FLAME

The quantized version of the model is less precise but has greater recall than the other models. This appears to be a trend in applying these optimizations. The precision tends to drop with pruning and quantization.

### 5.1.3 Zhang Et. Al Dataset

When observing the effectiveness of YOLO, it appears to be able to discern the location of distant smokestacks well. It is worth noting that this data is synthetically generated, so it may not have as much noise and lead to some overfitting. However, it can detect the smoke even when the background helps to camouflage it. This is prevalent in the top left cell and suggests that YOLO is rather robust.

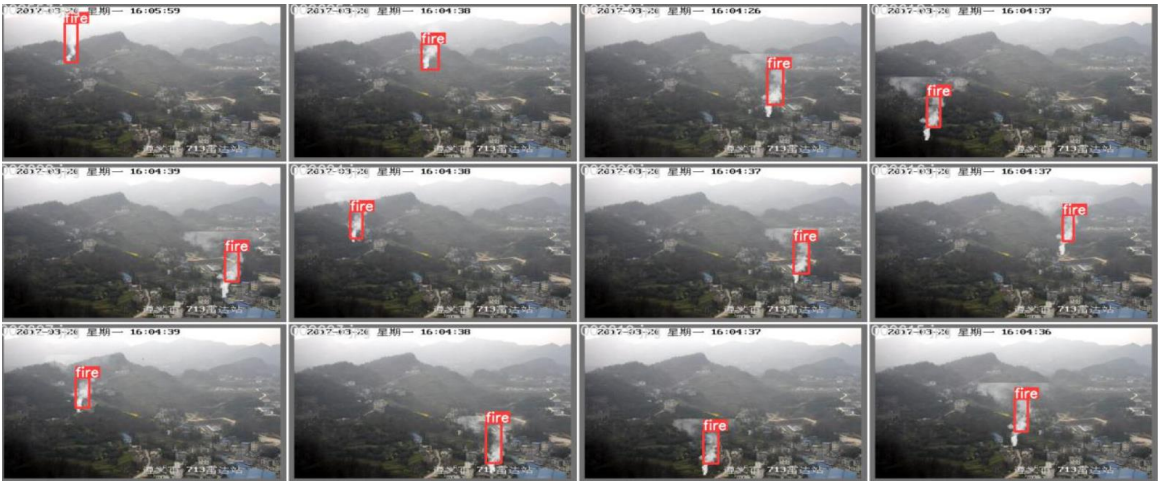

Figure 16:Mosaic of YOLO Fire Detection with Zhang et al. dataset

## 5.1.3.1 Base Model

| Comparison of YOLO With Zhang et Al | | | |
|---|---|---|---|
| Models | Precision | Recall | mAP |
| Base Model | 70.9% | 70.2% | 13.92% |
| Pruned Model | 56.6% | 70.2% | 9.77% |
| Quantized Model | 69.2% | 75.0% | 10.09% |
| Reports max value assiociated | | | |

Figure 17:Comparison of YOLO With Zhang et al

The mAP of the base model is 14%. It is important to recognize that this metric is dependent on the overlapping of bounding boxes. Since the contours of the smoke are not defined well, it is likely throwing off the metric.

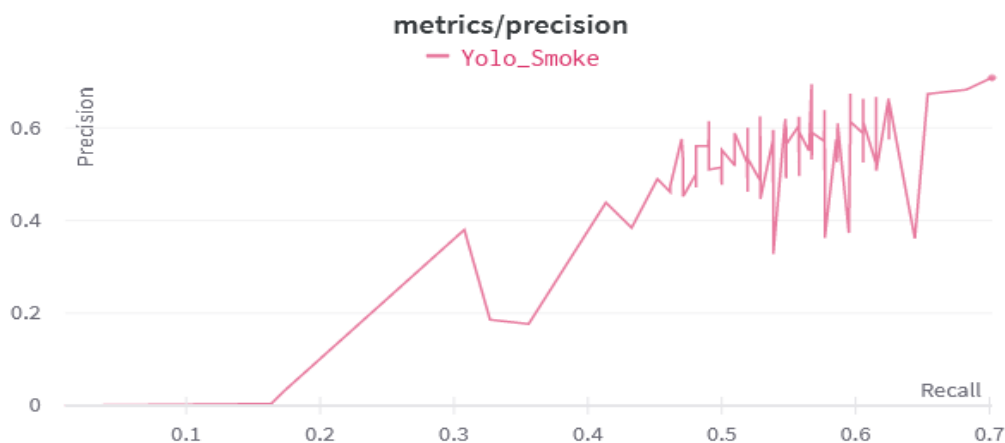

Figure 18:Recall/Precision YOLO Base with Zhang et al.

It reaches 70% precision and recall suggesting that the model is rather balanced. This could be reflecting the artificial nature of the data.

### 5.1.3.2 Pruned Model

The pruned model has 10% mAP which is less than the base. It reaches a max precision of 56% and a 70% in recall. The model essentially becomes weaker in detecting false positives. This could be problematic when there are similar objects in frame. For instance, morning steam is commonly confused with smoke. This could create a significant problem for deployment.

### 5.1.3 .3 Quantized Model

The quantized model has 10% mAP. It reaches a max 69% precision and 75% recall. In terms of the precision-recall graph, the quantized model performs slightly better.

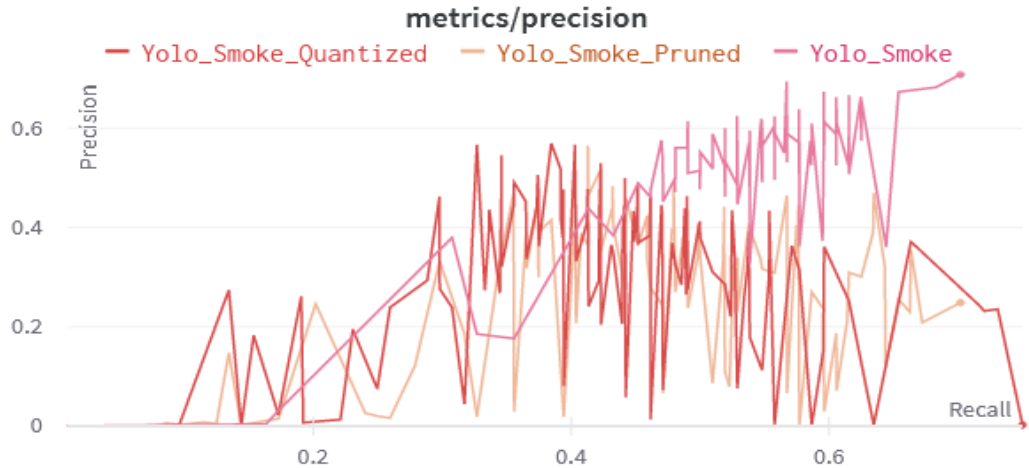

Figure 19: Recall/Precision of YOLO across models with Zhang et al.

# 5.2 Xception Evaluation

Xception is a CNN model that typifies a significant portion of the field. It is based on residual blocks and to promote locality.

## 5.2.1 BowFire Dataset

The BowFire dataset is the smallest dataset reviewed. It provides a diverse luminosity of flames that is valuable, but it may not be large enough and may encourage overfitting.

### Comparison of Xception With BowFire

| Models | TP | FP | TN | FN | Precision | Recall | AUC | Accuacy |
|---|---|---|---|---|---|---|---|---|
| Base Model | 22 | 24 | 0 | 0 | 47% | 100% | 0.59 | 47.0% |
| Pruned Model | 5 | 1 | 23 | 17 | 83% | 22% | 0.72 | 60.0% |
| Quantized Model | 0 | 0 | 24 | 2 | 0% | 0% | 0.74 | 52.0% |

Figure 20: Comparison of Xception With BowFire

## 5.2.1.1 Base Model

The model is evaluated on the test set that is held from the training and contains 20% of images in the dataset. The model is overly skewed to determine if there is a fire. It seems unable to handle negative cases. This may be seen in the difference between recall and precision and the overall accuracy.

There is an additional metric is included in this reporting. The AUC (Area Under the Curve) is a measure of specificity. An AUC of .5 is essentially random. Thus, the model may be better than a coin flip.

## 5.2.1.2 Pruned Model

The pruned version of the model seems to be more performant. This is likely due to the model removing some overfitting due to the smaller nature of the data. The model makes a trade-off between precision and recall. There are fewer false positives for false negatives. The general accuracy of the model is higher. This is the general goal of the model. In addition, the AUC is significantly higher: around .75 is considered fair with 1 being perfect specificity. Thus, the model is more useful.

## 5.2.1.3 Quantized Model

The quantized model is essentially the inverse case of the original model. It just assigns negative classes to all the input images. It is interesting how after pruning, the gains in precision are lost entirely. The model is still better than a coin flip in terms of accuracy, but this is likely the result of the data being very slightly imbalanced.

## 5.2.2 FLAME Dataset

The Xception had transferred learning applied to this dataset. In Shamsoshoara et al, 76% accuracy is reported in the validation set. The authors reported 3888 true negatives, 799 false positives, 1248 false negatives, and 2680 true positives. The precision reported is 77% and recall reported is 68%

| Comparison of Xception With FLAME | | | | | | | | |
|---|---|---|---|---|---|---|---|---|
| Models | TP | FP | TN | FN | Precision | Recall | AUC | Accuacy |
| Base Model | 2230 | 778 | 4359 | 1250 | 74% | 64% | 0.85 | 76.0% |
| Pruned Model | 1583 | 547 | 4590 | 1897 | 74% | 45% | 0.86 | 71.0% |
| Quantized Model | 0 | 0 | 5137 | 3480 | 0% | 0% | 0.83 | 59.0% |

Figure 21:Comparison of Xception With FLAME

## 5.2.2.1 Base Model

In general, the authors' results were able to be produced with slight variation. The model has strong precision, recall, and accuracy. The AUC is also suggestive of a healthy classifier. Accuracy is reproduced in terms of matching previous research.

## 5.2.2.2 Pruned Model

The model losses some performance overall. Most notably, there is a 19% decrease in recall. There are simply more false negatives that are being classified. In terms of classification of positive cases, the change is more minor. There is a 5% drop in accuracy, but the model still remains robust.  There is no drop in precision despite pruning.

## 5.2.2.3 Quantized Model

The effect of quantizing the model seems to affect models in a similar fashion. It appears to shift the weights to make classification favor negative cases. This is the case when quantizing BowFire. Quantizing simply diminishes precision and recall. It is beginning to appear that the trade-off with quantizing is too significant. With BowFire, it appeared to be an issue with the smaller dataset. The FLAME dataset should be significantly large, so the loss of positive classifications is rather troubling.

### 5.2.3 Zhang Et. Al Dataset

This dataset uses artificial smoke to synthetically populate images of a rural area. The concern when training the model is that the model may overfit. There is a lack of natural variation in the dataset and only includes positive cases. The negative cases from the FLAME dataset were used to augment the dataset for binary classification and to add additional variance. Photos of mountains were mostly included.

| Comparison of Xception With Zhang et al | | | | | | | | |
|---|---|---|---|---|---|---|---|---|
| Models | TP | FP | TN | FN | Precision | Recall | AUC | Accuacy |
| Base Model | 2592 | 0 | 2552 | 280 | 100% | 90% | 1.0 | 94% |
| Pruned Model | 2597 | 0 | 2552 | 275 | 100% | 90% | 1.0 | 94% |
| Quantized Model | 270 | 0 | 2552 | 2602 | 100% | 9% | 0.9 | 52% |

Figure 22:Comparison of Xception With Zhang et al

### 5.2.3.1 Base Model

The model seems to be able to classify on the dataset well. The precision is concerning. It suggests the model may not be generalizable.

### 5.2.3.2 Pruned Model

Pruning seems to have had little impact on the performance of the model. There are more true positives and fewer false negatives. The literature suggests that pruning could help address issues with overfitting, so it may make the model slightly more robust.

### 5.2.3.3 Quantized Model

Quantization has the same effect as in the other models. It overwhelmingly encourages classification of negative cases. The effect is not as severe in the other models as some true positives are still correctly classified. The drop in accuracy though is of great concern. Quantizing the model reduced accuracy by 42%. This leaves the once rather accurate models marginally better than a coin flip.

# 5.3 Visual Transformer

Due to the nature of transformers, it is necessary to use transfer learning. Imagenet's weights are used for the transfer learning process. This is the typical implementation.

## 5.3.1 BowFire Dataset

A generator is used based on the test set for evaluating the transformer. Since the dataset is balanced, it classifies an equal amount of classifications. Thus, accuracy is equal to recall and precision.

**Comparison of Transformer With BowFire**

| Models | AUC | Accuracy |
|--------|-----|----------|
| Base Model | 99% | 95.0% |
| Pruned Model | 67% | 61.0% |
| Quantized Model | 67% | 62.0% |

Figure 23:Comparison of Transformer

### 5.3.1.1 Base Model

It appears that the model is doing well. It is rather accurate and balanced in terms of precision and recall. Also, the size of the dataset is less of a concern and leverages the variance of the images of BowFire through transfer learning.

## 5.3.1.2 Pruned Model

It is surprising that pruning is so significant. Only the dense layers are pruned, and attention layers should not have been affected. It seems like pruning may undo some of the learning done in the transfer process. The architecture may not be as robust as CNN are to pruning effects. There is a 34% drop.

## 5.3.1.3 Quantized Model

Quantizing the model seems to not have had a significant effect in terms of accuracy or other performance metrics. Pruning may be more damaging, since tuning the model with new data would naturally create weaker weights than what is already present. Quantizing might preserve some of these newer relationships that were recently learned by the model.

## 5.3.2 FLAME Dataset

The FLAME dataset is from Shamsoshoara et al. The accuracy the author reported is 76%. The precision reported is 77% and recall reported is 68%

### Comparison of Transformer With FLAME

| Models | AUC | Accuracy |
|---|---|---|
| Base Model | 89% | 80.0% |
| Pruned Model | 78% | 75.0% |
| Quantized Model | 78% | 75.0% |

Figure 24:Comparison of Transformer With FLAME

### 5.3.2.1 Base Model

It appears that using a transformer outperforms Xception. It is more accurate and maintains a balance with the other metrics. This suggests an advance of methods that is encouraging although marginal.

### 5.3.2.2 Pruned Model

When pruning the model, there is a slight drop in accuracy and other performance metrics. It is interesting that the drop is not nearly as significant in other models. This may be a factor of the quality of the data. If the weights that are more associated with detecting fires are not being pruned away, then it may be more effectively learning the particular features for detecting fires.

## 5.3.2.3 Quantized Model

Quantizing the model did not have a significant impact. Only the dense layers in the tail of the model were quantized which may explain the lack of impact. It is recommended to quantize later in the model and to avoid layers closer to attention mechanisms. This may explain why there was no drop in metrics.

### 5.3.3 Zhang et. Al Dataset

When reviewing these results, it is important to consider that the data is created in a syntenic fashion and suggests that data may overfit in certain conditions. In Zhang et al, the authors were conducting object detection with the smoke but reported detection with 99% accuracy depending on the variety of smoke.

## 5.3.3.1 Base Model

Although the articulation of the problem differs, the model seems comparable to the R-CNN that is used in Zhang et al. The main difference is that a bounding box is not used as the task here is simply image classification.

| Comparison of Transformer With Zhang et al | | |
|---|---|---|
| **Models** | **AUC** | **Accuracy** |
| Base Model | 100.0% | 100% |
| Pruned Model | 92.0% | 88% |
| Quantized Model | 92.0% | 88% |

Figure 25:Comparison of Transformer With Zhang et al

## 5.3.3.2 Pruned Model

There is a significant penalty for pruning the model. But it is not as steep as the case with the FLAME dataset. It is interesting how this penalty fluctuates across datasets.

## 5.3.3.3 Quantized Model

There also appears to be no penalty associated with quantizing the model. Quantizing is applied to the tail of the model to avoid attention mechanisms and to avoid adversely affecting model predictions. It is surprising that there is no change, but this may be a factor of pruning removing the weights leaving less impact to be done during quantization.

## 5.4 Ensemble Approach

| Precision | Recall | AUC | Accuracy |
|-----------|--------|-----|----------|
| 69% | 69% | .79 | 69% |

Figure 26 Result of Ensemble

Additional experiment with the models occurred through applying an ensemble approach. The results of the FLAME transformer and the transformer trained from Zhang et al were averaged together. The motivation here is that the transformer trained on Zhang et al.'s data would be more effective at capturing the smoke present in the FLAME dataset. This appears to not have been the case. It might be that since the data from Zhang et al is presented in a rural environment and not in a forested area, the resulting ensemble was more conflated. This appears to have the result.

# 6. Conclusion

The goal of this analysis has been to review the data that is available in the literature and to provide an inspection into the techniques to aid practitioners in refining their models To that end, this discussion will provide recommendations based from the experience gained from training and experimenting with the previous models and analysis.

## 6.1 Recommendations

The first concern when attempting to train a model is to determine the nature of the data available. If the data is limited, as is the case when attempting to train on BowFire, the results will often overfit. When the data is limited but of merit to train on, pruning should be applied to the model to enable it to be more generalizable. There is also the option of generating synthetic data. This is the path Zhang et al follows, but it shares many of the same problems but to a lesser degree. Given the two options, it would likely be best to use more data augmentation techniques to leverage the dataset.

Transfer learning is another solution to limited data. The caveat here is that it seems strongly recommended to use a CNN architecture over other architectures. When trying to optimize CNN through pruning, there was little if any penalty associated with accuracy losses [19]. However, if another architecture is desirable, then it is necessary that a trade-off with accuracy must be accepted. To mitigate losses through pruning, it appears that carefully discerned hyperparameters may guard against substantial performance loss. When training YOLO through the Neural Magic API, hyperparameters were already derived. This likely contributed in part to the performance of the models despite optimizations applied.

In terms of model selection, transformers are outperforming CNN architectures. But, they are more difficult to optimize. Transfer learning is required for them to have an advantage. This is potentially problematic as more pre-processing and other choices are being made before the researcher can start to make choices of their own and enables dependencies to build.  For instance, many models including the transformer reviewed, use Imagenet as the basis of their weights. Imagenet has been noted to have been mislabeled. This mislabeling likely has proliferated into the modeling here, and there is no recourse. The need for transfer learning also makes it more difficult to optimize transformers with pruning. This is problematic when attempting transfer learning with smaller datasets. For instance, the 34% loss with BowFire is notable. A sufficiently large dataset is needed to support transfer learning for a transformer. Conversely, CNN architectures do not handle quantization well. Quantization undermines precision and recall. For instance, quantizing the Xception model with the FLAME dataset left recall and precision at 0. The transformers did not seem as affected by quantization. They may have already reached the floor after pruning though. It may make sense that quantization should be applied to transformer models without pruning to maintain the weaker more recently learned relationships.

There are a series of decisions that must be made when doing applied machine learning. If the data is insufficient, you may simply accept the data will overfit and attempt to correct it with pruning and other techniques. Or, transfer learning may be applied which is the more common approach. But one must accept the liability of using weights from another model. This should not be taken lightly considering the critical environment and risk involved. Considering deployment, CNN appears more conducive to optimization while transformers have an edge in performance. If the data available is sufficiently large, then transformers will likely be the best choice. When training the FLAME data on the transformer, even when it was quantized and pruned, it performed at the Xception benchmark. Although the quantization only applied to the dense layers in the tail, the transformers demonstrated robustness.

## 6.2 Research Limitation

There are two limitations in the work. The first concerns training YOLO and having the optimizing applied.  The Neural Magic API is used. This provided an ease of implementation and is used in production settings to better reflect the needs of the intended consumer of this research. However, while training with their API, it is suggested that you use their hyperparameters. Thus, when observing the results of the comparison of YOLO, the combined effect of the optimization with hyperparameters is observed. Tuning hyperparameters for YOLO would have been out of scope for this research requiring thousands of GPU hours. This difference in tuning dampens the observable effect, but it is the difference in pruning and quantization that is significant with the hyperparameters held constant between the two. Additionally, the Neural Magic API uses an early stopping mechanism and cuts off the training epoch for pruning prematurely. This slightly weakens the comparisons between versions of the model.

As for transformers, training on balanced data resulted in having all accuracy metrics be equal. This makes it more difficult to discern trade-offs while prioritizing the AUC of the classifier. It is a more justifiable trade-off, but it weakens the ability to explore the relationship between recall and precision.

## 6.3 Future Work

Moving forward from this analysis, it seems prudent to attempt further experiments with ensemble methods. In the FLAME dataset, the authors collected an auxiliary dataset they did not use in their paper using a FLIR infrared camera. Training two transformers and attempting to ensemble them might lead to better predictions as thermal information is not being used in the training yet and affords another dimension to leverage.

In terms of more general advancement, reviewing how transformers react to other optimizations will be necessary for the field to adapt to the new architecture. Additionally, there are newer models that have been designed recently that intermix CNN and transformers. These are called ConVit architectures and may react less strongly to pruning [35]. The behavior of these architectures is not well studied and provides an avenue of future study to determine the most adaptable models.

# 8. Appendix

## Regarding Artefacts

Given the size of the weights of the models and the data, the vast majority of the artefacts are stored in Google Drive.[1] In the artefact directory, the base models for Xception and YOLO are submitted with a "READ ME." Zhang et al is labeled 'Smoke', and Xception is labeled 'UAV.'

# Artefact List

1. YOLO Model Trained on BowFire
2. YOLO Model Trained on FLAME
3. YOLO Model Trained on Zhang et al
4. Pruned YOLO Model Trained on BowFire
5. Pruned YOLO Model Trained on FLAME
6. Pruned YOLO Model Trained on Zhang et al
7. Quantized YOLO Model Trained on BowFire
8. Quantized YOLO Model Trained on FLAME
9. Quantized YOLO Model Trained on Zhang et al
10. Xception Model Trained on BowFire
11. Xception Model Trained on FLAME
12. Xception Model Trained on Zhang et al
13. Pruned Xception Model Trained on BowFire
14. Pruned Xception Model Trained on FLAME
15. Pruned Xception Model Trained on Zhang et al
16. Quantized Xception Model Trained on BowFire
17. Quantized Xception Model Trained on FLAME
18. Quantized Xception Model Trained on Zhang et al
19. Transformer Trained on Bowfire
20. Transformer Trained on Flame
21. Transformer Trained on Zhang et al.
22. Pruned Transformer Trained on Bowfire
23. Pruned Transformer Trained on Flame
24. Pruned Transformer Trained on Zhang et al.
25. Quantized Transformer Trained on Bowfire
26. Quantized Transformer Trained on Flame
27. Quantized Transformer Trained on Zhang et al.
28. Ensemble Transformer

---

[1] https://drive.google.com/drive/folders/1LjJVPWb7-KFKQYVsfQpsKr1_5eiPAIWX?usp=sharing